\documentclass[runningheads]{llncs}
\usepackage{graphicx}
\usepackage{booktabs}
\usepackage{url}
\usepackage{tikz}
\usepackage{comment}
\usepackage{amsmath,amssymb}
\usepackage{dsfont}
\usepackage{colortbl}
\usepackage{epsfig}
\usepackage[inline]{enumitem}
\usepackage[skip=0pt]{caption}
\usepackage{soul}
\usepackage[utf8]{inputenc}
\usepackage{listings}
\usepackage[T1]{fontenc}
\usepackage{booktabs}
\usepackage{amsfonts}
\usepackage{bm}
\usepackage{nicefrac}
\usepackage{microtype}
\usepackage{amsmath}
\usepackage{algpseudocode}
\usepackage{algorithm}
\usepackage{numprint}
\usepackage{siunitx}
\usepackage{etoolbox}
\usepackage{cancel}
\newcommand{\ubold}{\fontseries{b}\selectfont}
\robustify\ubold
\usepackage{bbold}
\usepackage{wrapfig}
\usepackage[skip=2pt,font=scriptsize,labelfont=scriptsize]{subcaption}
\usepackage{pifont}
\usepackage{lineno}
\usepackage{array,multirow,graphicx}
\usepackage{booktabs} 
\usepackage{longtable} 
\usepackage[para]{footmisc}
\usepackage{pgfplots}
\usepackage{mathtools}
\pgfplotsset{compat=1.18}
\usepackage[accsupp]{axessibility}  

\usepackage{xcolor}
\usepackage[pagebackref,breaklinks,colorlinks=true,linkcolor=blue,citecolor=blue]{hyperref}

\usepackage{orcidlink}

\begin{document}

\title{LOGOS: Language-guided Oriented Object Detection in Aerial Scenes}

\titlerunning{LOGOS: Language-guided Oriented Object Detection in Aerial Scenes}

\author{
Trong-Thuan Nguyen\orcidlink{0000-0001-7729-2927}\inst{1,2}
\and
Minh-Triet Tran\orcidlink{0000-0003-3046-3041}\inst{1,2}}

\authorrunning{Trong-Thuan Nguyen and Minh-Triet Tran}

\institute{University of Science, VNU-HCM, Vietnam 
\and
Vietnam National University, Ho Chi Minh City, Vietnam
\email{ntthuan@selab.hcmus.edu.vn}, \email{tmtriet@fit.hcmus.edu.vn}
}

\maketitle

\begin{abstract}
Object detection in geospatial scenes, such as satellite and aerial imagery, poses significant challenges due to the varying orientations and densities of objects, as well as the complex backgrounds inherent to remote sensing imagery. Traditional methods for oriented object detection have struggled to address issues such as angular discontinuity, fixed query sizes, and inefficiencies in handling sparse or cluttered scenes. In this paper, we propose LOGOS, a novel transformer-based approach that leverages textual prompts to guide the detection of oriented objects in aerial scenes. In particular, our proposed approach incorporates prompt-modulated content queries to dynamically adjust the model’s focus based on the provided text, thereby improving object detection accuracy in complex environments. Empirically, extensive experiments on the DOTA dataset demonstrate that LOGOS outperforms existing state-of-the-art methods, particularly in densely packed and rotated object scenarios. Our approach offers a significant step forward in improving the robustness and scalability of oriented object detection in remote sensing applications.
\end{abstract}

\section{Introduction}\label{sec:intro}
\begin{wrapfigure}[13]{r}{0.6\textwidth}
  \centering
  \vspace{-21mm}
  \includegraphics[width=1\linewidth]{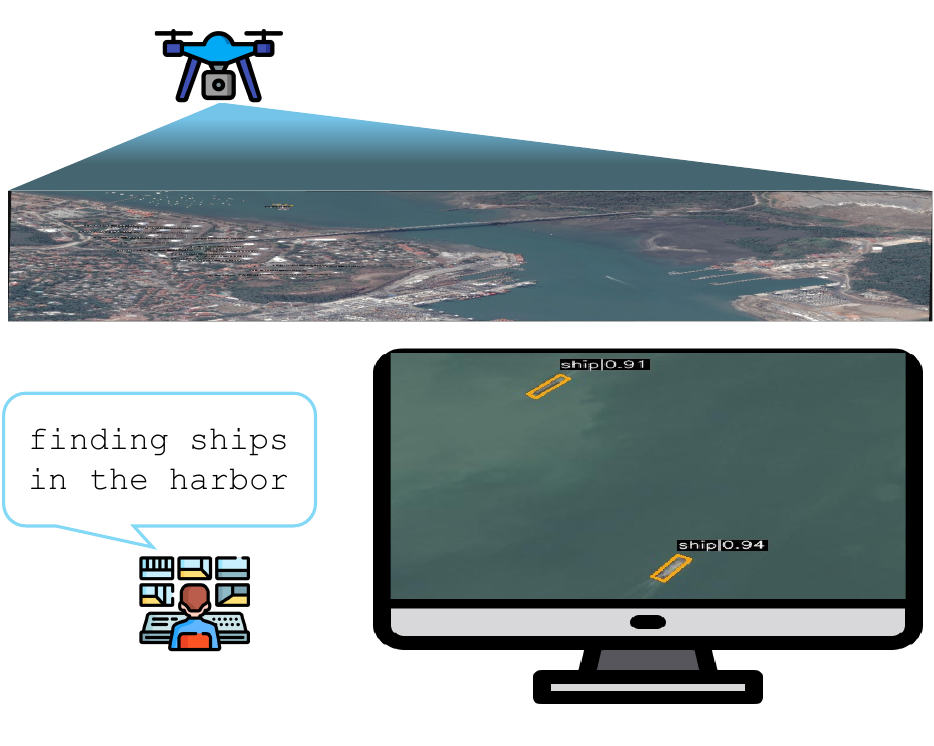}
  \caption{Illustration of our approach, where textual prompts are injected to guide the detection.}
  \label{fig:demo_live}
\end{wrapfigure}
Object detection in optical remote sensing images is a crucial task across applications such as urban planning, disaster management, agricultural monitoring, and military surveillance. In particular, remote sensing images, such as those obtained from satellites or drones, provide a bird's-eye view (BEV) of the Earth's surface, offering rich information about the spatial distribution of objects. In addition, these oriented objects vary significantly in size, shape, and orientation, making the detection task particularly challenging. Traditionally, object detection in images has focused on axis-aligned boxes, but in remote sensing, the orientation of objects often plays a significant role, as objects like buildings, vehicles, and ships can appear at arbitrary angles~\cite{ding2021object}~\cite{hu2023emo2}~\cite{xu2023dynamic}.

Despite significant advancements, existing methods for oriented object detection in remote sensing images face several limitations. Most early methods extend conventional object detection techniques to handle rotated bounding boxes. However, these oriented object detection methods typically use a classification-based angle prediction, which struggles with the inherent angular periodicity of rotation, where orientations at $-\pi/2$ and $\pi/2$ are equivalent but are inconsistently predicted~\cite {dai2022ao2}. Furthermore, square-like bounding boxes often introduce orientation ambiguity, leading to confusion during training and negatively affecting model performance~\cite{yang2021r3det}. Additionally, various methods rely on a fixed number of object queries, which can lead to inefficiencies when detecting images with fewer objects, as they risk excessive queries and poor convergence, leading to overfitting~\cite{carion2020end}\cite{zhu2020deformable}.

To address these challenges, there is growing interest in leveraging transformer-based models, such as the Detection Transformer (DETR)~\cite{carion2020end} and DINO~\cite{zhang2022dino}, for horizontal object detection. In particular, these models treat the object detection task as a set prediction problem, enabling a more flexible approach to handle various object densities and orientations~\cite{sun2021sparse}\cite{hu2023emo2}\cite{dai2022ao2}. However, while DETR-based models have shown promising results in image object detection, their direct application to remote sensing images often falls short due to the unique characteristics of these images, such as larger image sizes, complex background clutter, and significant variation in object density and orientation~\cite{ding2021object}. Specifically, the current methods for oriented object detection have yet to fully address the specific challenges posed by remote sensing benchmarks, particularly when integrating object-specific information, such as text, into the detection process.

The motivation behind this work is to overcome these limitations by proposing a novel language-guided approach to oriented object detection. Specifically, the core idea is to condition the detection process on a textual prompt that specifies the objects to be detected. In particular, this enables the model to focus on relevant oriented objects, thereby improving detection accuracy. By directly incorporating language as a guiding mechanism, we can handle highly cluttered and complex scenes, where objects may be densely packed or obscured~\cite{dai2022ao2}\cite{zhang2022dino}. In addition, our approach aims to address orientation ambiguity and reduce prediction redundancy by using a more dynamic, flexible query-based framework. 

In this paper, we propose LOGOS (\textbf{\textit{l}}anguage-\textbf{\textit{g}}uided \textbf{\textit{o}}riented object detection in aerial \textbf{\textit{s}}cenes), a transformer-based method for oriented object detection that incorporates textual prompts to guide the detection process. Our approach uses a prompt-modulated query mechanism, where the input text prompt conditions the visual features, enabling the model to focus on specific object categories while handling their orientations. Through extensive experiments on the DOTA dataset~\cite{ding2021object}, we show that our method outperforms state-of-the-art methods at mAP and achieves significant improvements in oriented object detection.

The rest of the paper is organized as follows: Sec.~\ref{sec:related} reviews related work in the areas of oriented object detection and transformer-based models, particularly in the context of remote sensing images. Sec.~\ref{sec:formulation} describes the problem formulation and defines the key notations used in our approach. In Sec.~\ref{sec:method}, we detail the architecture of our proposed approach, including the prompt-modulated content queries and the text-aware cross-attention mechanism. In Sec.~\ref{sec:exp}, we present the experimental setup and provide quantitative and qualitative results, highlighting the performance of our proposed method in comparison to existing methods. Finally, in Sec.~\ref{sec:conclusion}, we conclude this paper and discuss future research directions.

\section{Related Work}\label{sec:related}
In this section, we first discuss the evolution of transformer-based object detection methods and then explore various methods for oriented object detection. Finally, we highlight the limitations of previous methods and position our approach.

\subsection{Detection Transformers}\label{rw:detr}
The advent of DETR-based methods has revolutionized object detection by leveraging the flexibility of transformers for modeling long-range dependencies in images. Early works, such as DETR~\cite{carion2020end}, Deformable DETR~\cite{zhu2020deformable}, and UPDETR~\cite{dai2021up}, have demonstrated state-of-the-art performance in natural scene object detection. Unlike traditional approaches, DETR eliminates the need for non-maximum suppression (NMS) during post-processing. Instead, it leverages self-attention mechanisms to learn dependencies between object queries and image features, allowing for the suppression of redundant detections directly within the model. In addition, the Hungarian loss is used to match each predicted object to a unique target or background, thereby optimizing the bipartite matching process.

To address limitations of DETR, Deformable DETR~\cite{zhu2020deformable} introduces sparse attention, focusing on key regions in the image, which significantly improves the model’s ability to detect smaller objects. UPDETR~\cite{dai2021up} further enhances performance by incorporating unsupervised learning to pre-train the encoder and decoder components, improving the initial feature representations and speeding up convergence during training. These advancements have collectively advanced DETR-based models, making them a cornerstone for modern object detection.

\subsection{Oriented Object Detection}\label{rw:obd}
Oriented object detection, where objects may appear at arbitrary orientations, poses a significant challenge compared to traditional axis-aligned object detection. Various methods have been proposed to address this problem, which can be broadly categorized into CNN-based and hybrid CNN-transformer-based approaches. CNN-based methods extend traditional object detection frameworks by incorporating rotated components, such as Oriented R-CNN~\cite{xie2021oriented}, which uses an oriented Region Proposal Network (RPN) to generate rotation-aware proposals. Similarly, RoI-Tran aligns features through the object’s angle to better match regions of interest (RoI) to their corresponding objects, improving accuracy for rotated instances. R$^3$Det~\cite{yang2021r3det} introduces a feature refinement module that aligns feature points with predicted objects, treating angle prediction as a classification task to mitigate boundary issues that often arise in rotated object detection.

Despite significant improvement in CNN-based methods, recent transformer-based models for oriented object detection remain relatively underexplored. Some recent efforts, such as EMO2-DETR~\cite{hu2023emo2} and AO2-DETR~\cite{dai2022ao2}, have explored leveraging transformers for this task. In particular, EMO2-DETR is built upon Deformable DETR~\cite{zhu2020deformable}, which adapts transformer-based architectures to improve rotated object detection and mitigate redundancy in scenes with significant object variations. Generally, this adaptation addresses the aforementioned challenge of relative redundancy in scenes with many objects of varying orientations.

\subsection{Discussion}
\subsubsection{Limitations of Previous Methods}
Previous oriented object detection methods, particularly those based on DETR-like models, face several limitations that our proposed method addresses. One key challenge lies in the fixed number of queries used by DETR-based models. While this design works well in natural image settings, it becomes inefficient when applied to remote sensing images with varying object densities and sizes. In particular, remote sensing images can contain fewer objects or highly cluttered backgrounds, leading to either unnecessary computational overhead from excessive queries or the failure to detect certain objects due to insufficient queries. Therefore, traditional methods such as DETR and Deformable DETR struggle to scale effectively to remote sensing images due to the large variety in oriented object densities and scales~\cite{carion2020end}\cite{zhu2020deformable}.

Moreover, in oriented object detection, boundary discontinuity and orientation ambiguity remain critical issues. In particular, the periodicity of angles, where angles such as $-\pi/2$ and $\pi/2$ are equivalent, often results in inconsistent predictions from models that use traditional angle prediction mechanisms. Additionally, this problem has been highlighted in prior work, such as Oriented R-CNN and $R^3$Det, where angle prediction inconsistencies arise due to the angular periodicity of rotations~\cite{xie2021oriented}\cite{yang2021r3det}. Additionally, models like Oriented R-CNN and $R^3$Det tend to use a predefined angular range (e.g., 0 to 90 degrees), which causes poor handling of objects at angles outside that range~\cite{yang2021r3det}. Accordingly, this restriction severely limits their performance in applications such as remote sensing, where objects can appear at arbitrary orientations. Furthermore, ambiguity in square-like bounding boxes, due to their rotational equivalence, often leads to higher regression losses and confusion during training, as discussed in previous work~\cite{yang2021r3det}\cite{xie2021oriented}.

\subsubsection{Advantages of Our Approach}
Our approach, termed LOGOS, incorporates prompt-modulated content queries, thereby mitigating the issue of fixed query size commonly seen in DETR models. Specifically, textual prompts guide the model to focus on relevant objects, enabling it to adapt to varying object densities and handle both sparse and densely packed scenes efficiently. In addition, this flexibility reduces unnecessary background queries and avoids the overfitting problems typically associated with fixed-query models, a limitation of previous object detection methods such as DETR~\cite{carion2020end} and Deformable DETR~\cite{zhu2020deformable}.

Additionally, our approach overcomes the limitations of earlier methods, such as Oriented R-CNN and $R^3$Det, which suffered from angular discontinuities and limited angular ranges~\cite{xie2021oriented}\cite{yang2021r3det}. This is particularly useful in remote sensing images, where objects can appear at arbitrary angles. Moreover, LOGO's ability to focus on specific object categories, as guided by the text prompt, enables better handling of square-like bounding boxes, reducing ambiguity and improving the precision of box predictions, which is a challenge faced by earlier models~\cite{yang2021r3det}\cite{xie2021oriented}.

\section{Problem Formulation}\label{sec:formulation}
In this work, we address the problem of oriented object detection in aerial scenes, aiming to predict oriented bounding boxes for objects in optical remote sensing images conditioned on a textual prompt. In particular, the challenge lies in predicting axis-aligned oriented bounding boxes at specific oriented angles. 
\begin{figure}[!t]
    \centering
    \includegraphics[width=1\linewidth]{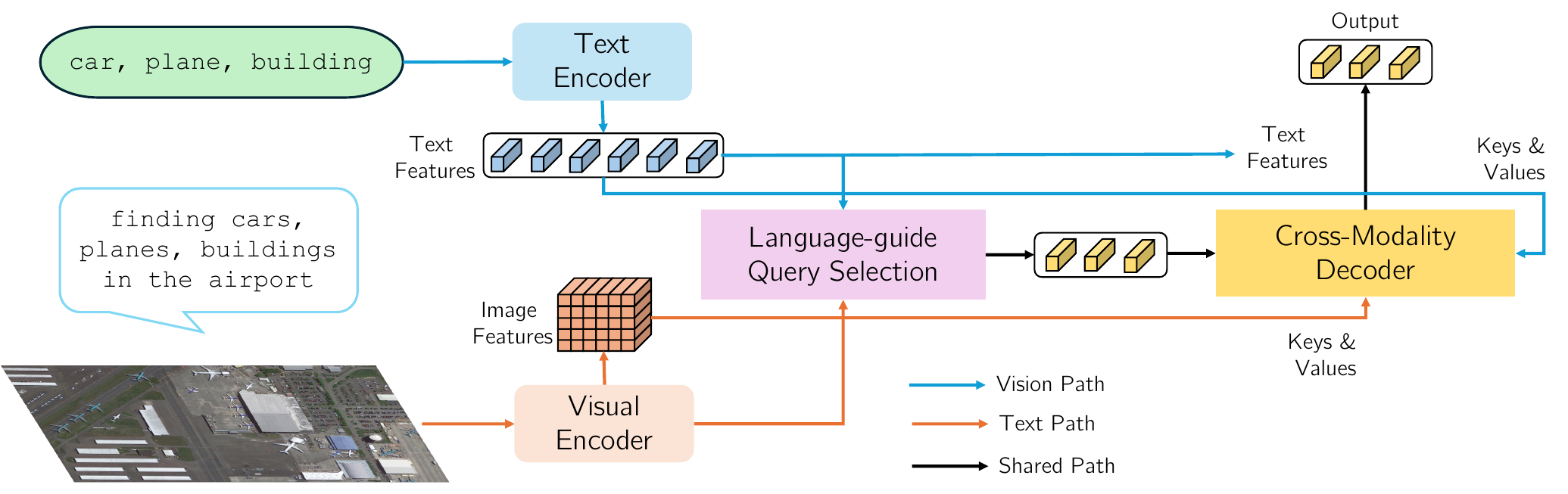}
    \vspace{\baselineskip}
    \caption{An overview of LOGOS, which is a language-guided approach for oriented object detection in aerial imagery. \textit{Best viewed in color and zoomed in.}}
    \label{fig:our_fw}
\end{figure}

\noindent\textbf{Notations}. Let $\mathbf{I} \in \mathbb{R}^{H \times W \times 3}$ be the RGB image and $\mathbf{P}$ the text prompt. Specifically, a DETR-style multi-scale encoder produces a sequence of visual tokens $\mathbf{enc}(\mathbf{I}) \in \mathbb{R}^{M \times D}$, where $M$ is the sequence length and $D$ is the token dimensionality. Simultaneously, the tokenized prompt is embedded as a matrix of text tokens, $\mathbf{emb}(\mathbf{P}) \in \mathbb{R}^{K \times D}$, where $K$ denotes the number of text tokens. Additionally, the decoder uses $Q$ learnable queries across $L$ layers, with a vocabulary size of $C$. 

Formally, an oriented bounding box is defined as 
    $\mathbf{b} = (x, y, w, h, \theta), \quad x \in [0, W), \, y \in [0, H), \, w, h > 0, \, \theta \in (-\pi, \pi]$,
where $x$ and $y$ represent the center coordinates, $w$ and $h$ the width and height of the box, and $\theta$ is the rotation angle.

\noindent\textbf{Task}. Given an image $\mathbf{I}$ and a prompt $\mathbf{P}$, we define the task as in Eqn.~\eqref{eqn:probabilistic_prediction}.
\begin{equation}\label{eqn:probabilistic_prediction}
    p(\mathcal{D} | \mathbf{I}, \mathbf{P}) = \prod_{q=1}^{Q} p(\mathbf{d}_q | \mathbf{I}, \mathbf{P}),
\end{equation}
where $p(\mathcal{D} | \mathbf{I}, \mathbf{P})$ is the probability distribution of the set of predicted bounding boxes $\mathcal{D}$, conditioned on the input image $\mathbf{I}$ and the prompt $\mathbf{P}$. Each individual bounding box prediction $\mathbf{d}_q = (x_q, y_q, w_q, h_q, \theta_q, s_q)$ is modeled as $p(\mathbf{d}_q | \mathbf{I}, \mathbf{P})$. 

\section{Our Proposed Approach}\label{sec:method}
In this section, we introduce LOGOS, which is a novel transformer-based approach that conditions object detection on a prompt. Specifically, we use prompt-modulated content queries that adjust the attention to focus on objects described in the prompt. An overview of our model architecture is shown in Fig.~\ref{fig:our_fw}.

\subsection{Prompt-Conditioned Oriented Detection}

\subsubsection{Prompt-Modulated Content Queries.}
In our approach, the model is an encoder-decoder architecture, where we use the encoder from DINO~\cite{zhang2022dino} to process the input image $\mathbf{I}$. The DINO encoder is leveraged to generate robust visual feature representations from the image, capturing both local and global object information. Specifically, the encoder directly produces visual tokens $\mathbf{enc}(\mathbf{I}) \in \mathbb{R}^{M \times D}$ from the image and text tokens $\mathbf{emb}(\mathbf{P}) \in \mathbb{R}^{K \times D}$ from the prompt, where $M$ and $K$ are the lengths and $D$ is the token dimensionality.

The decoder directly processes these visual and text tokens. Initially, it uses learnable content queries that are independent of the prompt. These content queries are then modulated by the input prompt using Feature-wise Linear Modulation (FiLM), which adjusts them based on the semantic information conveyed by the prompt. Formally, this modulation is defined as in Eqn.~\eqref{eqn:film}.
\begin{equation}\label{eqn:film}
    \mathbf{q}^{(0)}_{\mathrm{content}} = \mathrm{FiLM}(\mathbf{q}^{(0)}_{\mathrm{learn}}, \phi(\bar{\mathbf{p}})),
\end{equation}
where $\mathbf{q}^{(0)}_{\mathrm{learn}}$ represents the learnable content queries, and $\phi(\bar{\mathbf{p}})$ is a function that processes the pooled representation $\bar{\mathbf{p}}$ of the prompt. This modulation ensures that the queries are conditioned on the semantic content while retaining spatial.

\subsubsection{Text-aware Cross-Attention.}
After modulating the content queries, the decoder employs multi-head cross-attention (MHCA) to attend to both the visual and semantic features. The attention mechanism is formulated as in Eqn.~\ref{eqn:mhca}.
\begin{equation}\label{eqn:mhca}
    \mathrm{MHCA}(\mathbf{q}, \mathbf{K} = [ \mathbf{enc}(\mathbf{I}), \mathbf{emb}(\mathbf{P}) ], \mathbf{V} = [ \mathbf{enc}(\mathbf{I}), \mathbf{emb}(\mathbf{P}) ]),
\end{equation}
where $\mathbf{enc}(\mathbf{I})$ and $\mathbf{emb}(\mathbf{P})$ is the image and text embeddings. In particular, this mechanism enables queries to attend to both the image’s spatial features and the prompt’s semantic content, thereby improving object detection.

\subsection{Prediction Heads and Decoding}
After cross-attention, the decoder produces predictions for each query, including bounding-box parameters and class labels. These predictions are computed by prediction heads that perform bounding-box regression and object classification.

In particular, each query is passed through a feed-forward neural network (FFN) to regress the bounding-box parameters $(x, y, w, h, \sin \theta, \cos \theta)$. The angle $\theta$ is encoded using sine and cosine functions to avoid discontinuities in the angle space. A class mask is applied to the output logits to ensure that only relevant object classes are considered during both training and inference. Then, the class mask $\mathbf{M}$ is applied as defined in Eqn.~\ref{eqn:mask}, which filters out irrelevant classes.
\begin{equation}\label{eqn:mask}
\tilde{z}_c =
\begin{cases}
z_c, & \text{if } c \in S, \\
-\infty, & \text{if } c \notin S,
\end{cases}
\end{equation}
where $S$ is the set of relevant classes specified by the textual prompt $\mathbf{P}$.






\subsection{Loss Function}
Based on the detection transformer~\cite{carion2020end}, the total loss is a weighted combination of classification and bounding box regression losses as formally defined in Eqn.~\ref{eqn:train_loss}.
\begin{equation}\label{eqn:train_loss}
    \mathcal{L}_{\mathrm{train}} = \lambda_{\mathrm{cls}} \mathcal{L}_{\mathrm{cls}} + \lambda_{\mathrm{box}} \mathcal{L}_{\mathrm{box}},
\end{equation}
where the box loss $\mathcal{L}_{\mathrm{box}}$ incorporates L1 loss, GIoU for oriented bounding boxes, and smooth angle loss. Besides, the classification loss $\mathcal{L}_{\mathrm{cls}}$ is applied to the output logits, either through a multi-class head or a class-agnostic head.

Additionally, we utilize Hungarian matching to align predicted boxes with ground-truth boxes. The matching cost is computed as defined in Eqn.~\ref{eqn:matching_cost}.
\begin{equation}\label{eqn:matching_cost}
    \text{Cost}(\hat{\mathbf{d}}, \mathbf{g}) = \alpha_{\mathrm{cls}} \, \mathrm{FL}(\hat{p}_y, 1) + \alpha_{\mathrm{L1}} \, \lVert \hat{\mathbf{b}} - \mathbf{b} \rVert_1 + \alpha_{\mathrm{RotGIoU}} \left( 1 - \mathrm{RotIoU}(\hat{\mathbf{b}}, \mathbf{b}) \right),
\end{equation}
where $\hat{p}_y$ represents the predicted class probability for class $y$, and $\hat{\mathbf{b}}$ and $\mathbf{b}$ are the predicted and ground truth bounding boxes. Moreover, we leverage the contrastive denoising as an auxiliary loss to help the model recover from noisy positive queries while suppressing irrelevant negative queries as defined in Eqn.~\ref{eqn:cdn_loss}.
\begin{equation}\label{eqn:cdn_loss}
    \mathcal{L}_{\mathrm{CDN}} = \lambda_1 \, \mathcal{L}(Q^+, \mathrm{GT}) + \lambda_2 \, \mathcal{L}(Q^-, \varnothing),
\end{equation}
where $Q^+$ and $Q^-$ refer to positive and negative queries, respectively.

\subsubsection{Model Inference.} During inference, a set of candidate OBBs is generated. In particular, a relevance score threshold $s \geq \gamma$ is applied to retain only the most relevant predictions. Subsequently, non-maximum suppression (NMS) is performed to remove redundant boxes. The NMS computes the IoU between pairs of bounding boxes, discarding one if their IoU exceeds a threshold $\gamma_{\mathrm{oNMS}}$, ensuring that only the most accurate and non-redundant boxes are retained.

\section{Experimental Results}\label{sec:exp}
In this section, we first describe the datasets, followed by the metrics and implementation details. Then, we compare our method with state-of-the-art methods and provide quantitative and qualitative results to highlight its strengths.

\subsection{Implementation Details}\label{subsec:details}
\begin{table}[!t]
\centering
\vspace{\baselineskip}
\caption{Comparison with state-of-the-art methods on DOTA-v1.0. The results highlighted in \textcolor{red}{\textbf{red}} and \textcolor{blue}{\textbf{blue}} indicate the best and second-best performance.}
\vspace{2mm}
\label{tab:10}
\resizebox{\textwidth}{!}{%
\begin{tabular}{
  @{}l| c| S[table-format=2.2] S[table-format=2.2] S[table-format=2.2] S[table-format=2.2] S[table-format=2.2] S[table-format=2.2] S[table-format=2.2] S[table-format=2.2] S[table-format=2.2] S[table-format=2.2] S[table-format=2.2] S[table-format=2.2] S[table-format=2.2] S[table-format=2.2] S[table-format=2.2] | S[table-format=2.2]
  @{}
}
\toprule
{\textbf{Method}} & {\textbf{Backbone}} & {\textbf{Plane}} & {\textbf{BD}} & {\textbf{Bridge}} & {\textbf{GTF}} & {\textbf{SV}} & {\textbf{LV}} & {\textbf{Ship}} & {\textbf{TC}} & {\textbf{BC}} & {\textbf{ST}} & {\textbf{SBF}} & {\textbf{RA}} & {\textbf{Harbor}} & {\textbf{SP}} & {\textbf{HC}} & {\textbf{mAP}} \\ 
\midrule
YOLOv5m + PSCD~\cite{yu2023phase} & R50 & 89.86 & 86.02 & 54.94 & 62.02 & \textcolor{red}{\textbf{81.90}} & \textcolor{blue}{\textbf{85.48}} & 88.39 & 90.73 & \textcolor{blue}{\textbf{86.90}} & \textcolor{blue}{\textbf{88.82}} & 63.94 & 69.19 & 76.84 & \textcolor{blue}{\textbf{82.75}} & 63.24 & 50.35 \\
FR-O~\cite{ren2015faster} & R101 & 79.09 & 69.12 & 17.17 & 63.49 & 34.20 & 37.16 & 36.20 & 89.19 & 69.60 & 68.96 & 69.40 & 52.52 & 49.69 & 44.80 & 46.40 & 52.93 \\
IE-Net~\cite{lin2019ienet} & R101 & 80.20 & 64.54 & 39.82 & 32.07 & 49.71 & 65.01 & 52.58 & 81.45 & 44.66 & 78.51 & 46.54 & 56.73 & 64.40 & 64.24 & 36.75 & 57.14 \\
Rotated Repoints~\cite{yang2019reppoints} & R150 & 83.36 & 63.71 & 36.27 & 51.58 & 71.06 & 50.35 & 72.42 & 90.10 & 70.22 & 81.98 & 47.46 & 59.50 & 50.65 & 55.51 & 3.07 & 59.15 \\
Deformable DETR-O~\cite{zhu2020deformable} & R50 & 85.64 & 69.52 & 42.96 & 56.17 & 74.24 & 71.85 & 77.54 & 89.91 & 79.60 & 78.96 & 44.38 & 56.28 & 62.90 & 68.31 & 50.54 & 67.25 \\
RoI Transformer~\cite{ding2019learning} & R101 & 88.64 & 78.52 & 43.44 & 75.92 & 68.81 & 73.68 & 83.59 & 90.74 & 77.27 & 81.46 & 58.39 & 53.54 & 62.83 & 58.93 & 47.67 & 69.56 \\
$R^3$Det~\cite{yang2021learning}  & R50 & 89.29 & 75.21 & 45.41 & 69.24 & 75.54 & 72.89 & 79.29 & \textcolor{red}{\textbf{90.89}} & 81.02 & 83.25 & 58.81 & 63.15 & 63.43 & 62.21 & 37.41 & 69.80 \\
DAL~\cite{ming2021dynamic} & R101 & 88.68 & 76.55 & 45.08 & 66.80 & 67.00 & 76.76 & 79.74 & 90.84 & 79.54 & 78.45 & 57.71 & 62.27 & 69.05 & 73.14 & 60.11 & 71.44 \\
SCRDet~\cite{yang2019scrdet} & R101 & 89.98 & 80.65 & 52.09 & 68.36 & 68.36 & 60.32 & 72.41 & 90.85 & 87.94 & 86.86 & 65.02 & 66.68 & 66.25 & 68.24 & 65.21 & 72.61 \\
DRN~\cite{pan2020dynamic} & Hourglass-104 & 89.45 & 83.16 & 48.98 & 62.24 & 70.63 & 74.25 & 83.99 & 90.73 & 84.60 & 85.35 & 55.76 & 60.79 & 71.56 & 68.82 & 63.92 & 72.95 \\
$S^2A$-Net~\cite{han2021align} & R50 & 89.11 & 82.84 & 48.37 & 71.11 & 78.11 & 78.39 & 87.25 & 90.83 & 84.90 & 85.64 & 60.36 & 62.60 & 65.26 & 69.13 & 57.94 & 74.12 \\
SASM~\cite{hou2022shape} & R50 & 86.42 & 78.97 & 52.47 & 69.84 & 77.30 & 75.99 & 86.72 & 90.89 & 82.63 & 85.66 & 60.13 & 68.25 & 73.98 & 72.22 & 62.37 & 74.92 \\
Gliding Vertex~\cite{xu2020gliding} & R101 & 89.64 & 85.00 & 52.26 & 77.34 & 73.01 & 73.14 & 86.82 & 90.74 & 79.02 & 86.81 & 59.55 & 70.91 & 72.94 & 70.86 & 57.32 & 75.02 \\
CFA~\cite{guo2022convex} & R101 & 89.26 & 81.72 & 51.81 & 67.17 & 79.99 & 78.25 & 84.46 & 90.77 & 83.40 & 85.54 & 54.86 & 67.75 & 73.04 & 70.24 & 64.96 & 75.05 \\
DCFL~\cite{xu2023dynamic} & R50 & N/A & N/A & N/A & N/A & N/A & N/A & N/A & N/A & N/A & N/A & N/A & N/A & N/A & N/A & N/A & 75.35 \\
Oriented Repoints~\cite{li2022oriented} & R50 & 87.02 & 83.17 & 54.13 & 71.16 & 80.18 & 78.40 & 87.28 & 90.90 & 85.97 & 86.25 & 59.90 & 70.49 & 73.53 & 72.27 & 58.97 & 75.97 \\
CSL~\cite{yang2020arbitrary} & R152 & 90.25 & 85.53 & 54.64 & 75.31 & 70.44 & 73.51 & 77.62 & 90.84 & 86.15 & 86.69 & 69.60 & 68.04 & 73.83 & 71.10 & 68.93 & 76.17 \\
GWD~\cite{yang2021rethinking} & R152 & 86.96 & 83.88 & 54.36 & 77.53 & 74.41 & 68.48 & 80.34 & 86.62 & 83.41 & 85.55 & 73.47 & 67.77 & 72.57 & 75.76 & 73.40 & 76.30 \\
DAFNe~\cite{lang2021dafne} & R101 & 89.40 & \textcolor{red}{\textbf{86.27}} & 53.70 & 60.51 & \textcolor{red}{\textbf{82.04}} & 81.17 & 88.66 & 90.37 & 83.81 & 87.27 & 53.93 & 69.38 & 75.61 & 81.26 & 70.86 & 76.95 \\
RIDet~\cite{ming2021optimization} & R50 & 89.31 & 80.77 & 54.07 & 76.38 & 79.81 & 81.99 & \textcolor{red}{\textbf{89.13}} & 90.72 & 83.58 & 87.22 & 64.42 & 67.56 & 78.08 & 79.17 & 62.07 & 77.62 \\
KLD & R50 & 88.91 & 85.23 & 53.64 & \textcolor{red}{\textbf{81.23}} & 78.20 & 76.99 & 84.58 & 89.50 & 86.84 & 86.38 & 71.69 & 68.06 & 75.95 & 72.23 & \textcolor{red}{\textbf{75.42}} & 78.32 \\
EMO2-DETR~\cite{hu2023emo2} & Swin-T & 87.62 & 83.30 & 56.15 & 77.58 & 79.24 & 83.56 & 86.56 & 90.78 & 84.38 & 86.47 & 63.47 & 65.91 & 77.85 & 82.50 & 71.55 & 78.46 \\
DARDet~\cite{zhang2021dardet} & R50 & 89.08 & 84.30 & 56.64 & 77.83 & 81.10 & 83.39 & 88.46 & \textcolor{blue}{\textbf{90.88}} & 85.44 & 87.56 & 62.77 & 66.23 & 77.97 & 82.03 & 67.40 & 78.74 \\
$AO^2$-DETR~\cite{dai2022ao2} & R50 & 89.95 & 84.52 & 56.90 & 74.83 & 80.86 & 83.47 & 88.47 & 90.87 & 86.12 & 88.55 & 63.21 & 65.09 & 79.09 & \textcolor{red}{\textbf{82.88}} & 73.46 & 79.22 \\
ReDet\cite{han2021redet} & ReR50 & 88.81 & 82.48 & 60.83 & \textcolor{blue}{\textbf{80.82}} & 78.34 & \textcolor{red}{\textbf{86.06}} & 88.31 & 90.87 & \textcolor{red}{\textbf{88.77}} & 87.03 & 68.65 & 66.90 & 79.26 & 79.71 & \textcolor{blue}{\textbf{74.67}} & 80.10 \\
Oriented R-CNN~\cite{xie2021oriented}  & R50 & 89.84 & \textcolor{blue}{\textbf{85.43}} & \textcolor{red}{\textbf{61.09}} & 79.82 & 79.71 & 85.35 & \textcolor{blue}{\textbf{88.82}} & \textcolor{blue}{\textbf{90.88}} & 86.68 & 87.73 & \textcolor{blue}{\textbf{72.21}} & \textcolor{blue}{\textbf{70.80}} & \textcolor{blue}{\textbf{82.42}} & 78.18 & 74.11 & \textcolor{blue}{\textbf{80.87}} \\
LOGOS (Ours) & R50 & \textcolor{red}{\textbf{95.23}} & 84.87 & \textcolor{blue}{\textbf{60.89}} & 79.75 & 79.33 & 82.24 & 85.81 & 87.42 & 77.60 & \textcolor{red}{\textbf{93.81}} & \textcolor{red}{\textbf{74.54}} & \textcolor{red}{\textbf{70.86}} & \textcolor{red}{\textbf{93.50}} & 81.57 & 72.35 & \textcolor{red}{\textbf{81.32}} \\
\bottomrule
\end{tabular}
}
\end{table}
\subsubsection{Datasets.} We evaluate on three versions of the DOTA~\cite{ding2021object} dataset, including DOTAv1.0, DOTAv1.5, and DOTAv2.0. In particular, these datasets comprise optical remote sensing images ranging from 800$\times$800 to 4000$\times$4000 pixels, with around 280k annotated instances across 15 categories. Additionally, these categories include movable objects (planes, ships, large vehicles, small vehicles, and helicopters) and new objects (roundabouts, harbors, swimming pools, etc.).

\subsubsection{Metric.} We evaluate model performance using the mean Average Precision (mAP), a standard metric for object detection tasks. Specifically, we report mAP@0.5:0.95, which computes the mean average precision across multiple intersection-over-union thresholds from 0.5 to 0.95, with a step size of 0.05. 

\subsubsection{Setup.} In our setup, the textual prompt is the full label category of an object present in an input image. For example, for an image containing planes and helicopters, the textual prompt will specify these categories, guiding the model to detect only the relevant objects. In particular, the predictions are conditioned on this text prompt, ensuring that the bounding boxes correspond to the described objects and accounting for both localization and orientation.

\subsubsection{Configuration.} The initial learning rate is set to $1 \times 10^{-4}$ and is adjusted using a learning rate scheduler that reduces the learning rate by a factor of 0.1 at the 11th, 20th, and 30th epochs for the 12, 24, and 36 epoch settings with ResNet50. The AdamW optimizer is used with a weight decay of $1 \times 10^{-4}$, and the model is trained on 1 $\times$ NVIDIA A100 GPU with a batch size of 8. To maintain computational parity with DN-DETR~\cite{li2022dn}, which utilizes 300 queries and 3 patterns, this study employs $300 \times 3 = 900$ decoder queries. The training process incorporates L1 loss and GIoU loss for bounding box regression, and focal loss with $\alpha = 0.25$ and $\gamma = 2$ for classification. The loss coefficients are set to 1.0 for the classification loss, 5.0 for the L1 loss, and 2.0 for the GIoU loss. To ensure stable convergence, early stopping is applied after a maximum of 40 epochs.

\subsection{Quantitative Results}\label{subsec:sota}

\begin{table}[!t]
\centering
\vspace{\baselineskip}
\caption{Comparison with state-of-the-art methods on DOTA-v1.5. The results highlighted in \textcolor{red}{\textbf{red}} and \textcolor{blue}{\textbf{blue}} indicate the best and second-best performance.}
\vspace{2mm}
\label{tab:v15}
\resizebox{\textwidth}{!}{%
\begin{tabular}{
  @{}l| c| S[table-format=2.2] S[table-format=2.2] S[table-format=2.2] S[table-format=2.2] S[table-format=2.2] S[table-format=2.2] S[table-format=2.2] S[table-format=2.2] S[table-format=2.2] S[table-format=2.2] S[table-format=2.2] S[table-format=2.2] S[table-format=2.2] S[table-format=2.2] S[table-format=2.2] S[table-format=2.2] | S[table-format=2.2]
  @{}
}
\toprule
{\textbf{Method}} & {\textbf{Backbone}} & {\textbf{Plane}} & {\textbf{BD}} & {\textbf{Bridge}} & {\textbf{GTF}} & {\textbf{SV}} & {\textbf{LV}} & {\textbf{Ship}} & {\textbf{TC}} & {\textbf{BC}} & {\textbf{ST}} & {\textbf{SBF}} & {\textbf{RA}} & {\textbf{Harbor}} & {\textbf{SP}} & {\textbf{HC}} & {\textbf{CC}} & {\textbf{mAP}} \\ 
\midrule
Deformable DETR-O~\cite{zhu2020deformable} & R50 & 70.06 & 60.01 & 40.21 & 52.51 & 37.75 & 71.05 & 78.71 & 89.69 & 74.39 & 67.52 & 33.21 & 54.96 & 58.28 & 65.99 & 44.6 & 3.84 & 56.42 \\
RetinaNet-O~\cite{lin2017focal} & R50 & 71.43 & 77.64 & 42.12 & 64.65 & 44.53 & 59.79 & 73.31 & \textcolor{blue}{\textbf{90.84}} & 76.02 & 59.96 & 46.95 & \textcolor{blue}{\textbf{69.24}} & 59.65 & 64.52 & 48.06 & 0.83 & 59.16 \\
EMO2-DETR~\cite{hu2023emo2} & R50 & 71.81 & 75.36 & 45.09 & 58.70 & 48.19 & 73.26 & 80.28 & 90.70 & 73.05 & 76.53 & 39.36 & 65.31 & 56.96 & 69.29 & 47.11 & 15.64 & 61.67 \\
FR-O~\cite{ren2015faster} & R50 & 71.89 & 77.64 & 44.45 & 59.87 & 51.28 & 68.98 & 79.37 & 90.78 & \textcolor{red}{\textbf{77.38}} & 67.50 & 47.75 & 69.72 & 61.22 & 65.25 & 60.45 & 1.52 & 62.00 \\
HTC-O~\cite{chen2019hybrid} & R50 & 76.84 & 73.51 & 49.90 & 57.80 & 51.31 & 71.34 & 79.75 & 90.46 & 74.21 & 66.07 & 46.21 & 70.61 & 63.07 & 64.46 & 57.81 & 9.42 & 62.67 \\
Mask R-CNN-O~\cite{he2017mask} & R50 & 76.84 & 73.51 & 49.90 & 57.80 & 51.31 & 71.34 & 79.75 & 90.46 & 74.21 & 66.07 & 46.21 & \textcolor{red}{\textbf{70.61}} & 63.07 & 64.46 & 57.81 & 9.42 & 62.67 \\
$AO^2$-DETR~\cite{dai2022ao2} & R50 & 79.55 & 78.14 & 42.41 & 61.23 & 55.34 & 74.50 & 79.57 & 90.64 & 74.76 & 77.58 & 53.56 & 66.91 & 58.56 & \textcolor{blue}{\textbf{73.11}} & \textcolor{red}{\textbf{69.64}} & \textcolor{blue}{\textbf{24.71}} & 66.26 \\
ReDet~\cite{han2021redet} & ReR50 & \textcolor{blue}{\textbf{79.20}} & \textcolor{red}{\textbf{82.81}} & \textcolor{blue}{\textbf{51.92}} & \textcolor{blue}{\textbf{71.41}} & \textcolor{blue}{\textbf{52.38}} & \textcolor{blue}{\textbf{75.73}} & 80.92 & 90.83 & 75.81 & 68.64 & 49.29 & 72.03 & \textcolor{blue}{\textbf{73.36}} & 70.55 & \textcolor{blue}{\textbf{63.33}} & 11.53 & 66.86 \\
DCFL~\cite{xu2023dynamic} & R50 & N/A &  N/A & N/A & N/A & 56.72 & N/A & 80.87 & N/A & N/A & 75.65 & N/A &  N/A & N/A & N/A & N/A & N/A & 67.37 \\
RT~\cite{ding2019learning} & R101 & 88.64 & \textcolor{blue}{\textbf{78.52}} & 43.44 & \textcolor{red}{\textbf{75.92}} & \textcolor{red}{\textbf{68.81}} & 73.68 & \textcolor{blue}{\textbf{83.59}} & 90.74 & \textcolor{blue}{\textbf{77.27}} & \textcolor{blue}{\textbf{81.46}} & \textcolor{red}{\textbf{58.39}} & 53.54 & 62.83 & 58.93 & 47.67 & N/A & \textcolor{blue}{\textbf{69.56}}\\
LOGOS (Ours) & R50 & \textcolor{red}{\textbf{93.24}} & 77.21 & \textcolor{red}{\textbf{52.58}} & 59.19 & 51.40 & \textcolor{red}{\textbf{82.00}} & 65.58 & \textcolor{red}{\textbf{94.60}} & 51.41 & \textcolor{red}{\textbf{85.57}} & \textcolor{blue}{\textbf{56.85}} & 63.31 & \textcolor{red}{\textbf{91.98}} & \textcolor{red}{\textbf{80.15}} & 51.11 & \textcolor{red}{\textbf{60.24}} & \textcolor{red}{\textbf{69.97}} \\
\bottomrule
\end{tabular}
}
\end{table}

Table~\ref{tab:10} illustrates the performance of various state-of-the-art methods for oriented bounding box detection on the DOTA-v1.0 dataset, where LOGOS (Ours)-based approach stands out as the top performer. By leveraging the DINO encoder and prompt conditioning, our approach achieves a mAP of 81.32\%, outperforming other models across categories such as Plane, Storage Tank, Harbor, and Roundabout. Especially, incorporating textual prompts enables the model to focus on specific object categories, thereby enhancing detection accuracy and localization, particularly in challenging or cluttered scenes. However, the model shows slightly lower performance in Ship and Baseball Court (BC) categories, where certain object types or scales may still pose challenges for our prompt-based approach.

In addition, our approach demonstrates its strength on DOTA-v1.5, as shown in Table~\ref{tab:v15}, achieving an overall mAP of 69.97\%. In categories such as Plane, Bridge, Large Vehicle (LV), and Harbor, the prompt conditioning helps the model prioritize relevant objects and filter out irrelevant regions, improving its detection capabilities. Notably, the model achieves a remarkable 94.60\% in detecting Traffic Cones (TC), a task that benefits significantly from the prompt’s ability to guide the model’s focus toward smaller, more specific objects. Significantly, this model's ability to refine attention on particular object types is a key advantage of our proposed approach, as evidenced by strong performance in categories such as Harbor (91.98\%) and Swimming Pool (80.15\%). While the model continues to perform well across most categories, lower performance on Ship and Baseball Court (BC) suggests that further refinement of prompt conditioning or more targeted training may be needed to improve detection in these specific cases.

Furthermore, on the DOTA-v2.0 dataset, Table~\ref{tab:v20} shows that the LOGOS (Ours)-based approach achieves an mAP of 66.04\%, maintaining its position as a top performer. The integration of prompt conditioning continues to enhance detection across categories such as Plane, Building, Bridge, Large Vehicles (LV), and Tennis Courts, where the model excels at handling diverse object types and sizes. The ability to inject textual prompts that focus on specific objects, such as helicopters and Ground Track Fields, enables the model to perform better in cluttered or densely packed environments, such as Harbors and LV, where overlapping or obscured objects might otherwise hinder detection. The versatility of the method is further demonstrated by its strong results across categories such as Small Vehicles (SV), Storage Tanks (ST), and Roundabouts, indicating its potential for wide deployment in urban planning, surveillance, and autonomous navigation. However, as with previous datasets, performance on Ship and Baseball Court (BC) could still benefit from additional prompt refinement or specialized training to handle these more complex object types more effectively.

\begin{table}[!t]
\centering
\vspace{\baselineskip}
\caption{Comparison with state-of-the-art methods on DOTA-v2.0. The results highlighted in \textcolor{red}{\textbf{red}} and \textcolor{blue}{\textbf{blue}} indicate the best and second-best performance.}
\vspace{2mm}
\label{tab:v20}
\resizebox{\textwidth}{!}{%
\begin{tabular}{
  @{}l| c| S[table-format=2.2]S[table-format=2.2] S[table-format=2.2] S[table-format=2.2] S[table-format=2.2] S[table-format=2.2] S[table-format=2.2] S[table-format=2.2] S[table-format=2.2] S[table-format=2.2] S[table-format=2.2] S[table-format=2.2] S[table-format=2.2] S[table-format=2.2] S[table-format=2.2] S[table-format=2.2] S[table-format=2.2] S[table-format=2.2] | S[table-format=2.2]
  @{}
}
\toprule
{\textbf{Method}} & {\textbf{Backbone}} & {\textbf{Plane}} & {\textbf{BD}} & {\textbf{Bridge}} & {\textbf{GTF}} & {\textbf{SV}} & {\textbf{LV}} & {\textbf{Ship}} & {\textbf{TC}} & {\textbf{BC}} & {\textbf{ST}} & {\textbf{SBF}} & {\textbf{RA}} & {\textbf{Harbor}} & {\textbf{SP}} & {\textbf{HC}} & {\textbf{CC}} & {\textbf{Air}} & {\textbf{Heli}} & {\textbf{mAP}} \\ 
\midrule
DAL~\cite{ming2021dynamic} & R50 & 71.23 & 38.36 & 38.60 & 45.24 & 35.42 & 43.75 & 56.04 & 70.84 & 50.87 & 56.63 & 20.28 & 46.53 & 33.49 & 47.29 & 12.15 & 0.81 & 25.77 & 0.00 & 38.52 \\
SASM~\cite{hou2022shape} & R50 & 70.30 & 40.62 & 37.01 & 59.03 & 40.21 & 45.46 & 44.60 & 78.58 & 49.34 & 60.73 & 29.89 & 46.57 & 42.95 & 48.31 & 28.13 & 1.82 & 76.37 & 0.74 & 44.53 \\
RetinaNet-O~\cite{lin2017focal} & R50 & 70.63 & 47.26 & 39.12 & 55.02 & 38.10 & 40.52 & 47.16 & 77.74 & 56.86 & 52.12 & 37.22 & 51.75 & 44.15 & 53.19 & 51.06 & 6.58 & 64.28 & 7.45 & 46.68 \\
$R^3$Det~\cite{yang2021learning} & R50 & 75.44 & 50.95 & 41.16 & 61.61 & 41.11 & 45.76 & 49.65 & 78.52 & 54.97 & 60.79 & 42.07 & 53.20 & 43.08 & 49.55 & 34.09 & 36.26 & 68.65 & 0.06 & 47.26 \\
FR-O~\cite{ren2015faster} & R50 & 71.61 & 47.20 & 39.28 & 58.70 & 35.55 & 48.88 & 51.51 & 78.97 & 58.36 & 58.55 & 36.11 & 51.73 & 43.57 & 55.33 & 57.07 & 3.51 & 52.94 & 2.79 & 47.31 \\
FCOS-O~\cite{detector2022fcos} & R50 & 74.84 & 47.53 & 40.83 & 57.41 & 43.89 & 47.72 & 55.66 & 78.61 & 57.86 & 63.00 & 38.02 & 52.38 & 41.91 & 53.24 & 40.22 & 7.15 & 65.51 & 7.42 & 48.51 \\
Oriented Rep~\cite{li2022oriented} & R50 & 73.02 & 46.68 & 42.37 & 63.05 & 47.06 & 50.28 & 58.64 & 78.84 & 57.12 & 66.77 & 35.21 & 50.76 & 48.77 & 51.62 & 34.23 & 6.17 & 64.66 & 5.87 & 48.95 \\
Mask R-CNN-O~\cite{he2017mask} & R50 & 76.20 & 49.91 & 41.61 & \textcolor{blue}{\textbf{60.00}} & 41.08 & 50.77 & 56.24 & 78.01 & 55.85 & 57.48 & 36.62 & 51.67 & 47.39 & 55.79 & 59.06 & 3.64 & 60.26 & 8.95 & 49.47 \\
ATSS-O~\cite{zhang2020bridging} & R50 & 77.46 & 49.55 & 42.12 & 62.61 & 45.15 & 48.40 & 51.70 & 78.43 & 59.33 & 62.65 & 39.18 & 52.43 & 42.92 & 53.98 & 42.70 & 5.91 & 67.09 & 10.68 & 49.57 \\
$S^2A$-Net~\cite{han2021align} & R50 & 77.84 & 51.31 & 43.72 & 62.59 & 47.51 & 50.58 & 57.86 & 80.73 & 59.11 & 65.32 & 36.43 & 52.60 & 45.36 & 52.46 & 40.12 & 0.00 & 62.81 & 11.11 & 49.86 \\

HTC-O~\cite{chen2019hybrid} & R50 & 77.69 & 47.25 & 41.15 & 60.71 & 41.77 & 52.79 & 58.87 & 78.74 & 55.22 & 58.49 & 38.57 & 52.48 & 49.58 & 56.18 & 54.09 & 4.20 & 66.38 & 11.92 & 50.34 \\
RoI Transformer~\cite{ding2019learning} & R50 & 71.81 & 48.39 & 45.88 & 64.02 & 42.09 & 54.39 & 59.92 & 82.70 & 63.29 & 58.71 & 41.04 & 52.82 & 53.32 & 56.18 & 57.94 & 25.71 & 63.72 & 8.70 & 52.81 \\
Oriented R-CNN~\cite{xie2021oriented} & R50 & 77.95 & 50.29 & 46.73 & 65.24 & 42.61 & 54.56 & 60.02 & 79.08 & 61.69 & 59.42 & 42.26 & 56.89 & 51.11 & 56.16 & 59.33 & 25.81 & 60.67 & 9.17 & 53.28 \\

DCFL~\cite{xu2023dynamic} & R50 & 78.30 & 53.03 & 44.24 & 60.17 & 48.56 & 55.42 & 58.66 & 78.29 & 60.89 & \textcolor{blue}{\textbf{65.93}} & 43.54 & 55.82 & 53.33 & 60.00 & 54.76 & 30.90 & 74.01 & \textcolor{red}{\textbf{15.60}} & 55.08 \\
DCFL~\cite{xu2023dynamic} & R101 & 79.49 & 55.97 & 50.15 & 61.59 & \textcolor{blue}{\textbf{49.00}} & \textcolor{blue}{\textbf{55.33}} & \textcolor{blue}{\textbf{59.31}} & \textcolor{blue}{\textbf{81.18}} & \textcolor{blue}{\textbf{66.52}} & 60.06 & \textcolor{blue}{\textbf{52.87}} & \textcolor{red}{\textbf{56.71}} & \textcolor{blue}{\textbf{57.83}} & 58.13 & \textcolor{blue}{\textbf{60.35}} & \textcolor{red}{\textbf{35.66}} & \textcolor{blue}{\textbf{78.65}} & 13.03 & \textcolor{blue}{\textbf{57.66}} \\
LOGOS (Ours) & R50 & \textcolor{red}{\textbf{91.70}} & \textcolor{red}{\textbf{63.88}} & \textcolor{red}{\textbf{57.35}} & \textcolor{red}{\textbf{64.37}} & \textcolor{red}{\textbf{52.88}} & \textcolor{red}{\textbf{79.85}} & \textcolor{red}{\textbf{62.64}} & \textcolor{red}{\textbf{92.75}} & \textcolor{blue}{\textbf{55.99}} & \textcolor{red}{\textbf{77.02}} & \textcolor{red}{\textbf{61.51}} & \textcolor{blue}{\textbf{59.70}} & \textcolor{red}{\textbf{88.73}} & \textcolor{blue}{\textbf{80.51}} & \textcolor{red}{\textbf{67.71}} & \textcolor{blue}{\textbf{30.02}} & \textcolor{red}{\textbf{89.15}} & \textcolor{blue}{\textbf{13.04}}  & \textcolor{red}{\textbf{66.04}} \\

\bottomrule
\end{tabular}
}
\end{table}

\subsection{Qualitative Results}
\begin{figure}[!t]
    \centering
    \includegraphics[width=1\linewidth]{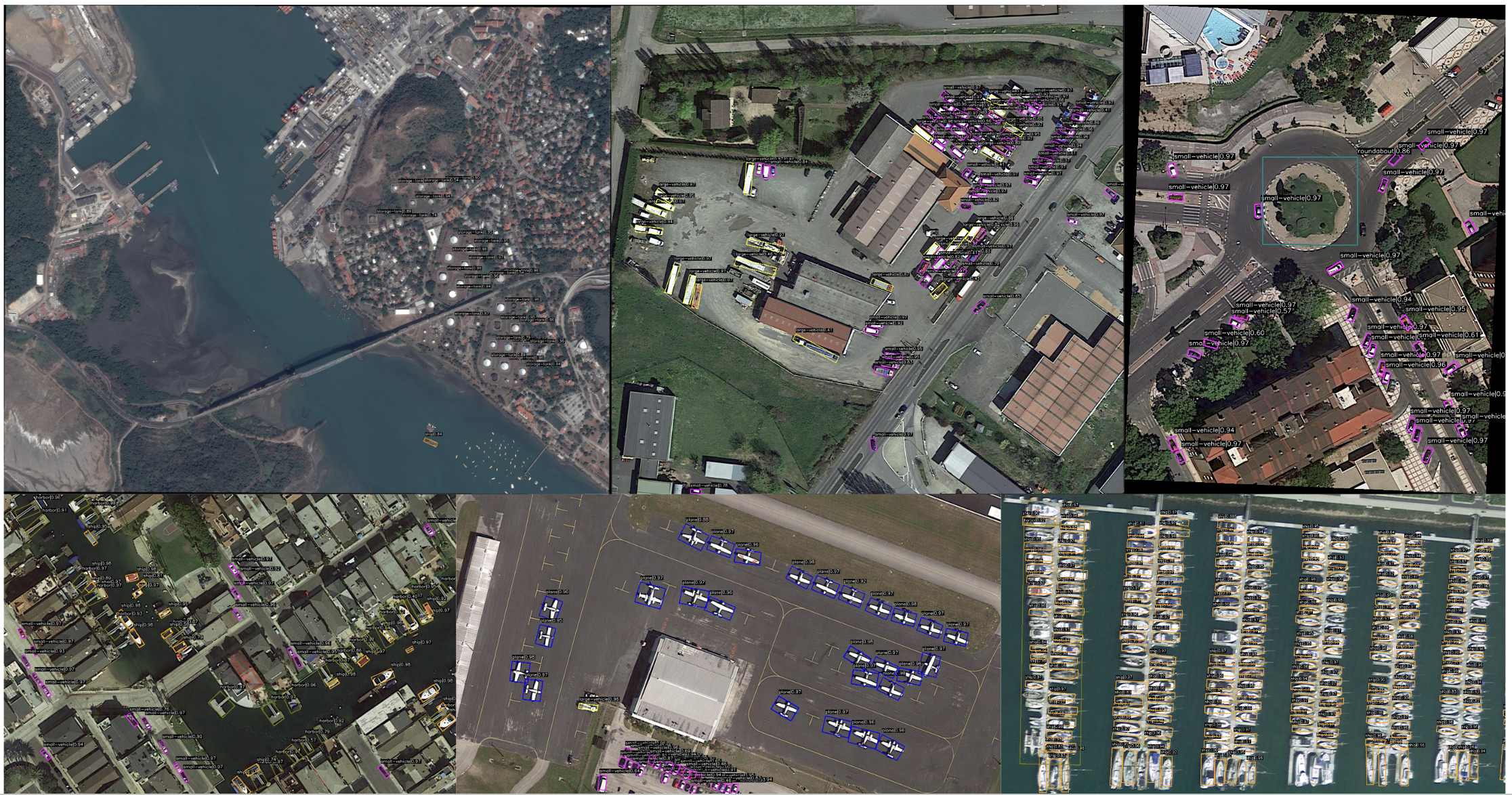}
    \vspace{\baselineskip}
    \caption{Qualitative results on DOTA. \textit{Best viewed in color and zoomed in.}}
    \label{fig:predictions}
\end{figure}

In Fig.~\ref{fig:predictions}, LOGOS detected oriented bounding boxes overlaid on optical remote sensing images. In particular, we observe that our model effectively detects various object types, even in densely packed, complex scenes. For example, the top-left image shows a harbor, where the method successfully identifies various ships despite the cluttered environment. In other regions (e.g., industrial and urban areas), the model excels at identifying specific objects, including small vehicles, buildings, and roundabouts, as evidenced by the purple bounding boxes. In addition, the precision of our proposed approach in handling diverse object scales, orientations, and environmental settings highlights its potential applications in areas such as urban planning, surveillance, and automated navigation.

\begin{figure}[!t]
    \centering
    \includegraphics[width=1\linewidth]{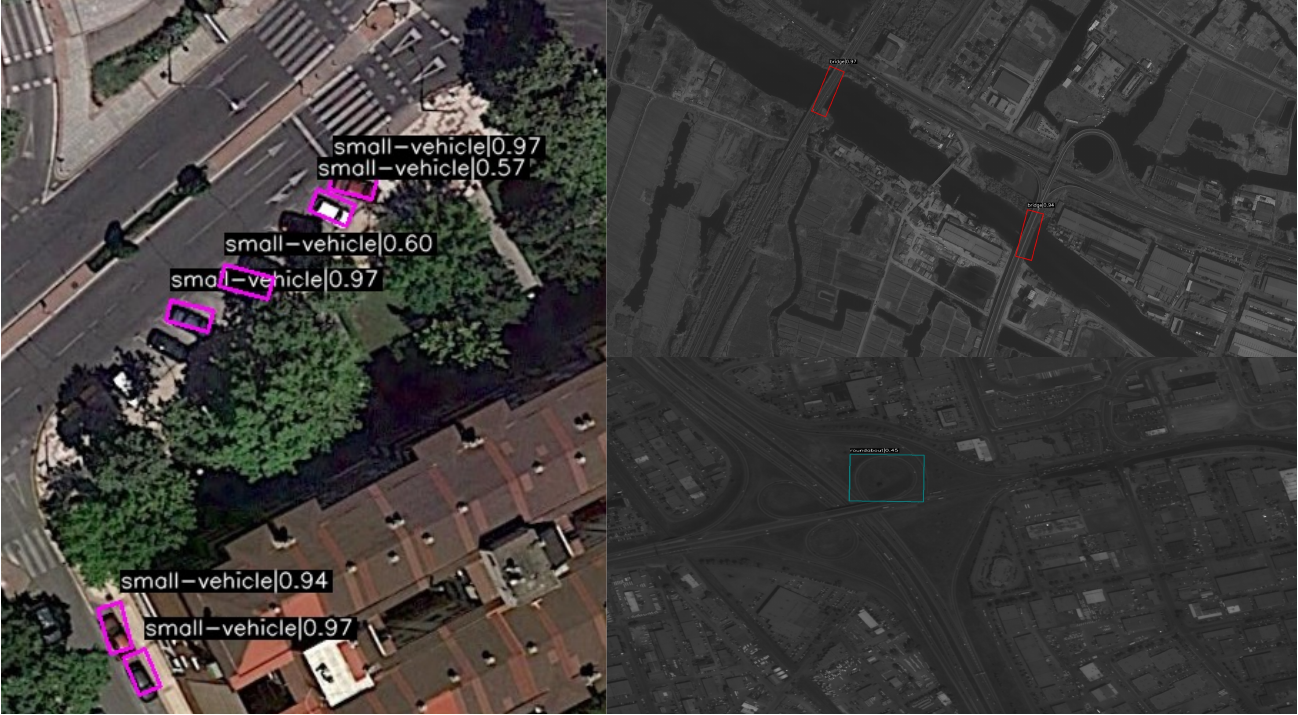}
    \vspace{\baselineskip}
    \caption{Illustration of failure cases. \textit{Best viewed in color and zoomed in.}}
    \label{fig:bad}
\end{figure}

Moreover, Fig.~\ref{fig:bad} shows failure cases in which the detector struggles in extreme conditions and in dense object environments. On the left, the model struggles to detect objects in a crowded scene with multiple small vehicles in close proximity, leading to incorrect or missed detections. The high object density and overlapping regions cause confusion, reducing the model's ability to accurately distinguish individual objects. On the right, the model struggles to detect objects in extreme conditions, such as when they are isolated or appear at unusual orientations, leading to low-confidence predictions and misalignment with the ground truth. These cases highlight the model's limitations in handling complex scenes with sparse or obstructed objects, suggesting areas for further improvement.

\section{Conclusion}\label{sec:conclusion}

In this paper, we have proposed LOGOS, a novel transformer-based approach for detecting rotated objects in remote sensing images. By conditioning the detection process on textual prompts, LOGOS effectively addresses challenges posed by varying object densities, orientations, and ambiguities in bounding box predictions. Empirically, we demonstrate that our proposed approach, using prompt-modulated content queries, outperforms state-of-the-art oriented object detection methods and improves accuracy in complex, cluttered environments.

Although our proposed approach has shown promising results, several avenues for future research remain to be explored. One potential direction is to enhance the model's ability to handle objects at extreme orientations, as the current approach may still face challenges with objects near $0$ or $180$ degrees, particularly in dense scenes. Additionally, integrating multimodal data, such as incorporating semantic information or high-resolution multispectral imagery, could further enhance the robustness and generalization. Finally, exploring the real-time applications, particularly in time-sensitive fields such as disaster management or surveillance, would require optimization and adaptation to handle large-scale datasets.

\vspace{5mm}
\textbf{Acknowledgment}. This research is funded by University of Science, VNU-HCM, under grant number T2025-158.

\newpage
\bibliographystyle{splncs04}
\bibliography{main}

\end{document}